\pdfoutput=1

\documentclass{article}

\usepackage[preprint, nonatbib]{neurips_2023}

\usepackage[utf8]{inputenc} 
\usepackage[T1]{fontenc}    
\usepackage{hyperref}       
\usepackage{url}            
\usepackage{booktabs}       
\usepackage{amsfonts}       
\usepackage{nicefrac}       
\usepackage{microtype}      
\usepackage{xcolor}         
\usepackage{graphicx} 
\usepackage{amsmath} 
\usepackage[numbers]{natbib}

\title{Bridging the Gap: Addressing Discrepancies in Diffusion Model Training for Classifier-Free Guidance}

\author{%
  Niket Patel\thanks{*Work done during internship at Swiss Data Science Center.} \\
  University of California, Los Angeles\\
  Los Angeles, CA 90095 \\
  \texttt{niketpatel@ucla.edu} \\
  \And
  {Luis Salamanca}\\
  Swiss Data Science Center, ETH Zürich,\\
  Zürich, 8092, Switzerland\\
  \texttt{luis.salamanca@sdsc.ethz.ch} \\
  \And
  {Luis Barba}\\
  Swiss Data Science Center, Paul Scherrer Institute,\\
  Villigen, 5232, Switzerland\\
  \texttt{luis.barba-flores@psi.ch} \\
}

\begin{document}

\maketitle

\begin{abstract}
Diffusion models have emerged as a pivotal advancement in generative models, setting new standards to the quality of the generated instances. In the current paper we aim to underscore a discrepancy between conventional training methods and the desired conditional sampling behavior of these models. While the prevalent classifier-free guidance technique works well, it's not without flaws. At higher values for the guidance scale parameter $w$, we often get out of distribution samples and mode collapse, whereas at lower values for $w$ we may not get the desired specificity. To address these challenges, we introduce an updated loss function that better aligns training objectives with sampling behaviors. Experimental validation with FID scores on CIFAR-10 elucidates our method's ability to produce higher quality samples with fewer sampling timesteps, and be more robust to the choice of guidance scale $w$. We also experiment with fine-tuning Stable Diffusion on the proposed loss, to provide early evidence that large diffusion models may also benefit from this refined loss function.
\end{abstract}

\section{Introduction}
Image generation has witnessed monumental shifts with the advent of diffusion models, which have set new gold standards for generation quality. However, the efficacy of conditional diffusion models can be heavily reliant on the guidance method employed. However, we find the loss function used during training may not directly align with the desired behavior of models during sampling. 

\subsection{Background on Diffusion Models}
Diffusion models function on the principle of score-based generative models \cite{song2021scorebased}. In practice, they are typically formulated as having some forward direction that adds Gaussian noise, denoted \(q(z_t|z_{t-1})\), and a parameterized reverse process that attempts to predict the noise added to the model, denoted \(p_\theta(z_{t-1}|z_t)\). In this paper we typically distinguish between the Denoising Diffusion Probablistic Model (DDPM) and the Denoising Diffusion Implicit Model (DDIM) sampling methods \cite{ho2020denoising, song2022denoising}. DDIM uses the same training procedure as DDPM, but enables to sample more efficiently with fewer timesteps.

\subsection{Guidance}

We usually apply guidance to intervene in the reverse process to create conditional samples. Classically, we can use classifier guidance on these score-based models by training a separate classifier network \cite{ho2022classifierfree}. For this approach, the sample is then modified during training by integrating the gradients from this classifier, as expressed in Eq.~\ref{cg}.

\begin{align}
   \tilde{\epsilon}_\theta(z_t, c ) = \epsilon_\theta(z_t, c) - w \sigma_t\nabla_{z_t} \log p_\phi(c|z_t)\label{cg}
\end{align}

Notice here that multiplying the second term by \(w\) means that we are effectively guiding our sample towards the distribution given by \(p_\phi(c|z_t)^w\), which for \(w>1\) should decrease sample diversity and increase sample class specificity \cite{ho2022classifierfree}.

However, due to the added complexity introduced by having to train a separate model, capable of classifying with different levels of noise, we see a shift towards classifier-free guidance \cite{ho2022classifierfree}. This method, which involves taking a linear combination of the outputs of both a conditional and an unconditional models to predict the noise during sampling, as shown in Eq.~\ref{cfg}, has set the precedent for many subsequent works. This approach is facilitated by a stochastic training regimen where conditional information is randomly dropped out, or replaced with a null token, during training, effectively creating a conditional and an unconditional model.

\begin{align}
\tilde{\epsilon}_\theta(z_t, c ) = (1+w)\epsilon_\theta(z_t, c) - w \epsilon_\theta(z_t, \emptyset)\label{cfg}
\end{align}

\subsection{Shortcomings in Classifier-Free Guidance}

Despite its widespread acceptance, classifier-free guidance presents some drawbacks. Most notably, larger values of the guidance parameter $w$ may lead to mode collapse and the generation of samples that deviate from expected distributions. We can see an example of this phenomena in Fig.~\ref{fig: toy}. Nevertheless, in practice, high values of $w$ are still often preferred to create samples that more accurately follow the conditioning information. The conditional samples provided in the Stable Diffusion paper use \(w=5\) and \(w=3\) \cite{Rombach_2022_CVPR}, and the primary text-to-video latent diffusion model in \cite{blattmann2023align} uses guidance values between \(w=8\) and \(w=13\). 

Previous literature notes similar problems when high values of \(w\) are used \cite{lin2023common}. This paper claims this occurs due to the scale of the resulting classifier-free guided noise becoming too high. They accordingly attempt to fix this by scaling to classifier-free guided sample to have the same variance as the purely conditional sample, then taking a convex combination of this re-scaled sample with a classical classifier-free guided sample.

\section{Methods}

Classically, classifier free-guidance (c.f.g) models are trained to predict the noise added to the image with the loss function introduced by \cite{ho2020denoising}, where conditioning $c$ is dropped out at random during the training process.

\begin{displaymath}
\mathcal{L}_{\text{standard}} = ||\epsilon - \epsilon_\theta(z_t, c)||^2_2
\end{displaymath}

However, during sampling the unconditional noise estimate is corrected with an extra term $w(\epsilon_\theta(z_t, c) - \epsilon_\theta(z_t, \emptyset))$, which guides the generation. This disparity between the training loss and the sampling process presents a disconnect: during sampling, a model that was trained to minimize \( \mathcal{L}_{\text{standard}} \) for both the conditional and unconditional case, doesn't accurately predict \(\epsilon\) when its prediction is modified by the classifier free guidance using Eq.~\ref{cfg}. If we train a model such that \(\epsilon_\theta(z_t, c )\) and \(\epsilon_\theta(z_t, \emptyset)\) are both good estimates of \(\epsilon\), then while a convex combination of the two may also be a good estimate of \(\epsilon\), this isn't necessarily the case for non-convex linear combinations of the two, which is what we have when we use higher values of the guidance scale parameter \(w\).

We root our rationale in the previous observation, and propose to incorporate the conditioning process more explicitly during the training process. Formally, our approach is notated by the following equation: 

\begin{displaymath}
\mathcal{L}_{\text{updated}} = || \epsilon - (1+w)\epsilon_\theta(z_t, c ) + w \epsilon_\theta(z_t, \emptyset) ||^2_2
\end{displaymath}

\begin{figure}
  \centering
  \includegraphics[width = 13.5 cm]{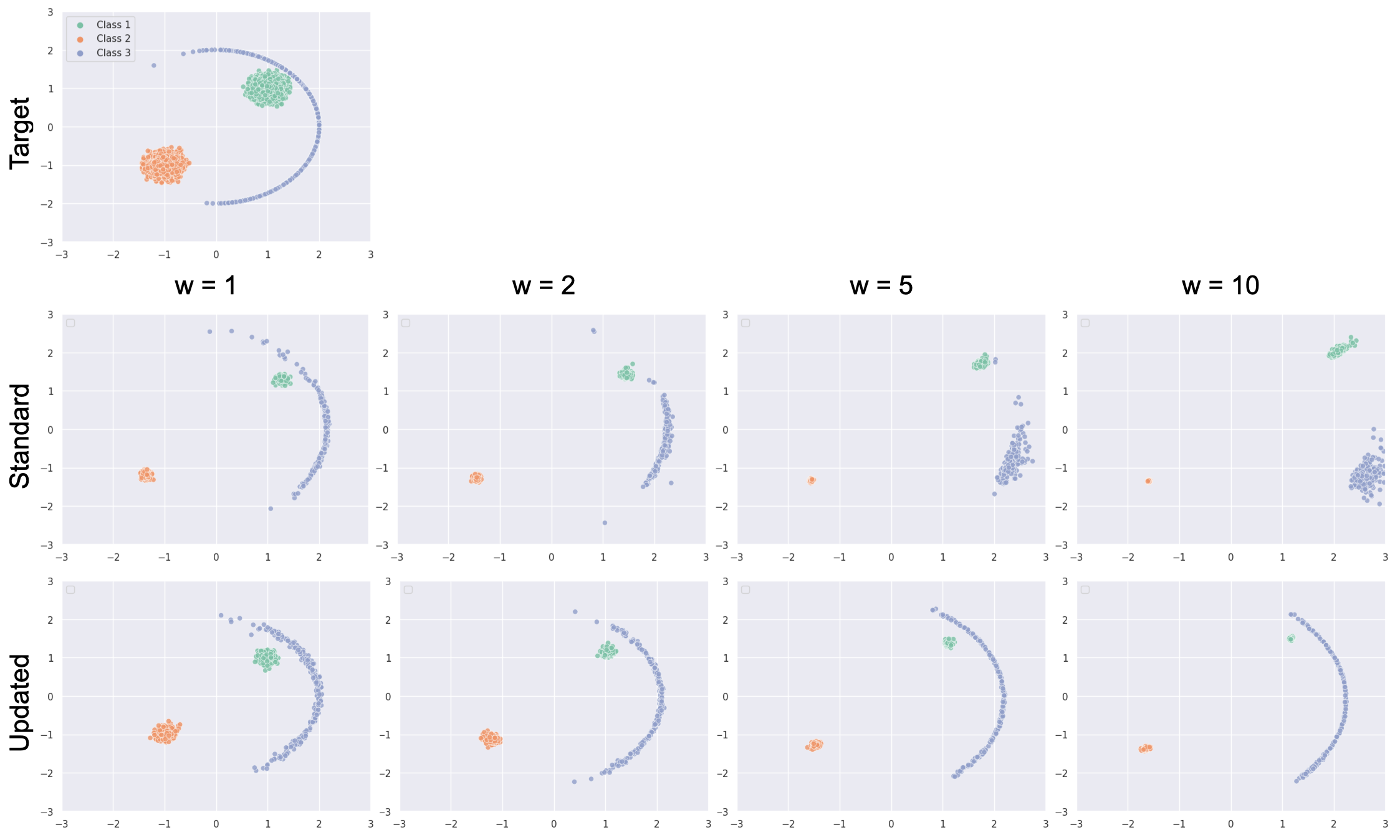}
  \caption{Comparison of standard classifier-free guidance (top row) to the updated loss function (bottom row) to target distribution (left-most column). The updated method more robustly represents accurate sampling guided from the distribution \(p(c|z_0)^w\) for different values of \(w\).}
  \label{fig: toy}
\end{figure}

In Fig.~\ref{fig: toy}, row ``Updated'', we observe how the generated distribution more closely matches the training distribution, presumably due to the training loss being better aligned with the sampling process. 

Given that the proposed loss somewhat differentiates the conditional and unconditional models, we don't expect that the unconditional model would produce high quality samples. It is also important to note that the proposed method comes with the downside that the effective computation required per batch is doubled, meaning in practice that, for some cases with limited VRAM, we may need to half the batch size. Nevertheless, the proposed loss could also be used in a fine-tuning step, which can potentially help alleviate the issues mentioned for models trained with the standard classifier-free guidance approach. 

We find empirically that this method performs well for broader ranges of \(w\) during sampling, as well as for values of \(w\) different from those used in training. We choose \(w = 1\) during training for most of the results presented in this paper. A figure detailing an ablation study on this training guidance parameter scale is presented in the appendix ~\ref{fig: ablation}. 

\section{Experiments}

\subsection{CIFAR-10 Benchmarking}

\begin{figure}
  \centering
  \includegraphics[height = 5.4 cm]{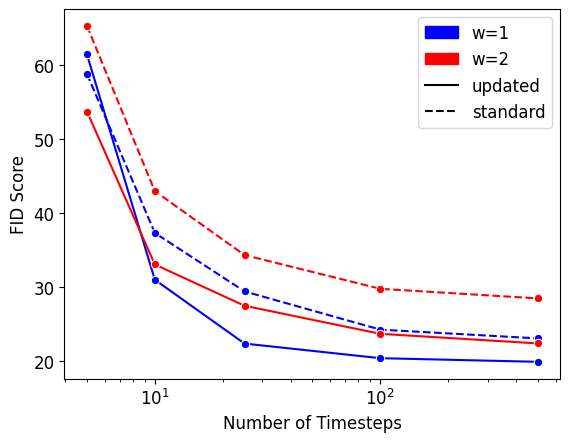}
  \caption{FID score with respect to number of DDIM sampling timesteps on two individually trained models on the CIFAR-10 Dataset.}
  \label{fig: cifar}
\end{figure}

To showcase the potential of the proposed loss, we first train two models on the CIFAR-10 dataset \cite{cifar} from scratch, one with the updated loss and the other with the standard c.f.g method. Accounting for the increased computational demand of our updated loss, and to perform a fair comparison, batch sizes were adjusted accordingly for the standard c.f.g. model. We used Frechet Inception Distance (FID) \cite{heusel2018gans} as our chosen metric for performance evaluation. The quantitative results in Fig.~\ref{fig: cifar} show a consistent performance improvement across varying \(w\) values and sampling steps, using a DDIM sampler \cite{song2022denoising}. Notably, the model trained with our updated loss retains its quality for fewer sampling steps, leading to an improved sampling efficiency. Over all timesteps tested in ~\ref{fig: cifar}, we see an average 15.62\% gain over the standard approach.

\begin{figure}
  \centering
  \includegraphics[width = 14 cm]{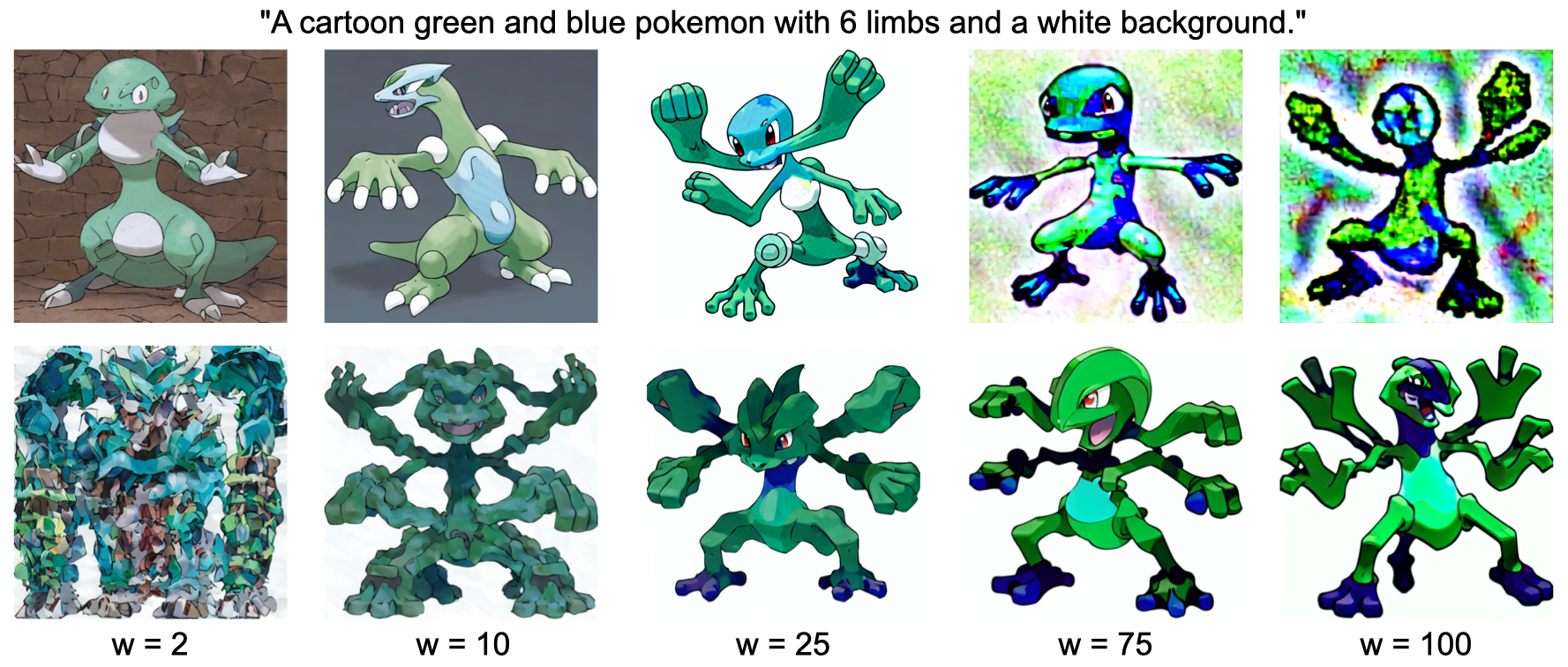}
  \caption{Results of fine-tuning Stable Diffusion on the Pokemon dataset. The top row is the model fine-tuned with the standard loss, while the bottom row is fine-tuned with our updated loss function.}
  \label{fig: pokemon}
\end{figure}

\subsection{Fine-Tuning Existing Models}
To further evaluate our approach, we fine-tuned the Stable Diffusion v1.4 model \cite{Rombach_2022_CVPR} using our updated loss on the Pokemon text-to-image dataset \cite{pinkney2022pokemon}. Qualitative assessments of generation quality are showcased in Fig.~\ref{fig: pokemon}. We find that our model performs better at higher values of $w$, and follows more precisely the prompt throughout the range of guidance values. We do note a failure case here for \(w = 2\), which could be due to the likely aforementioned poor performance of the unconditional model as a generator. More samples are provided in the appendix Fig.~\ref{fig: pokemon2}. 

\section{Discussion}

Our experiments highlight the potential of our updated loss function for training and fine-tuning diffusion models. By directly addressing the limitations of classifier-free guidance, we open the door to more reliable, robust, and faster image generation. Significantly, the enhanced efficiency of our model can allow for faster more accurate sampling, an improvement with critical implications for applications demanding real-time responses.

A tangible limitation of our approach is its impact on the maximum allowable batch size during training. Our method essentially halves the permissible batch size, posing potential challenges for setups constrained by memory and computational resources. We also loose the ability to generate samples unconditionally, so we have then a trade-off between generality of the model and specificity of conditional generation.

Further research could work towards deploying our enhanced loss function in more scenarios, particularly with larger-scale models and bigger datasets. Moreover, we envision an opportunity to further investigate and refine the proposed loss for fine-tuning, and assess its potential as a default fine-tuning step to improve established large diffusion models.

{\small
\bibliographystyle{plain}

}

\newpage
\section*{Appendix}
\subsection*{Ablation on guidance parameter used in training}

\begin{figure}[ht]
  \centering
  \includegraphics[width = 14 cm]{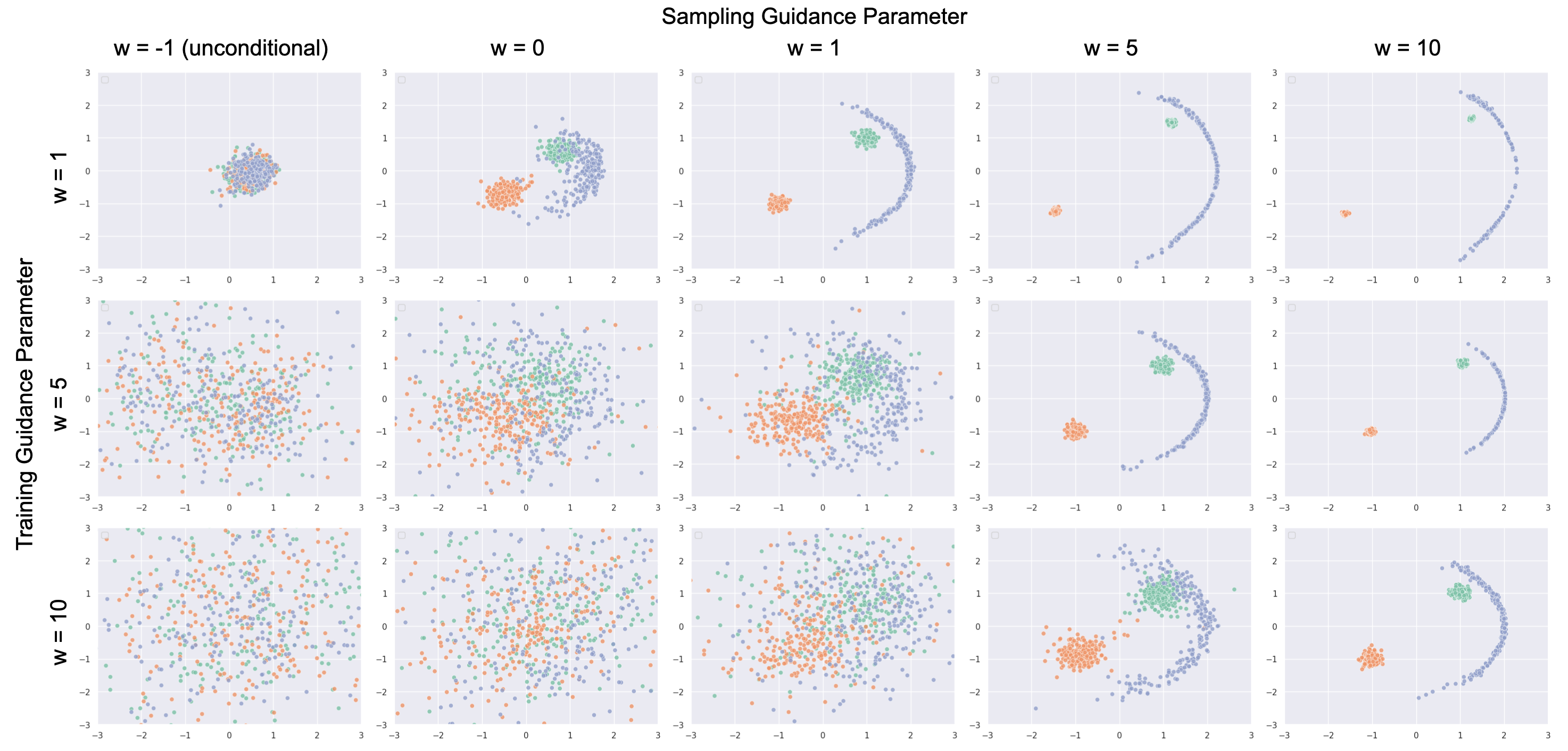}
  \caption{Ablation study of guidance parameter used during training (rows) versus the guidance parameter used during sampling (columns).}\label{fig: ablation}
\end{figure}

We show in Fig.~\ref{fig: ablation} the effects of using various guidance parameters in training (using the same target distribution as in Fig.~\ref{fig: toy}). We observe that we best match the target distribution when the sampling \(w\) matches that used in training. It follows that, to some extent, samples from a model trained with \(w_\text{train}\), could be related to the model trained with \(w = 1\), by \( w_\text{relative} \approx \frac{w_\text{sample}}{w_\text{train}}\)

To ensure that we can get high quality generation at various values for \(w\) and evaluate effectively against the standard c.f.g. model, we choose \(w = 1\) for most of the experiments in this paper. 

\clearpage
\subsection*{Stable Diffusion samples}

\begin{figure}[ht]
  \centering
  \includegraphics[width = 14 cm]{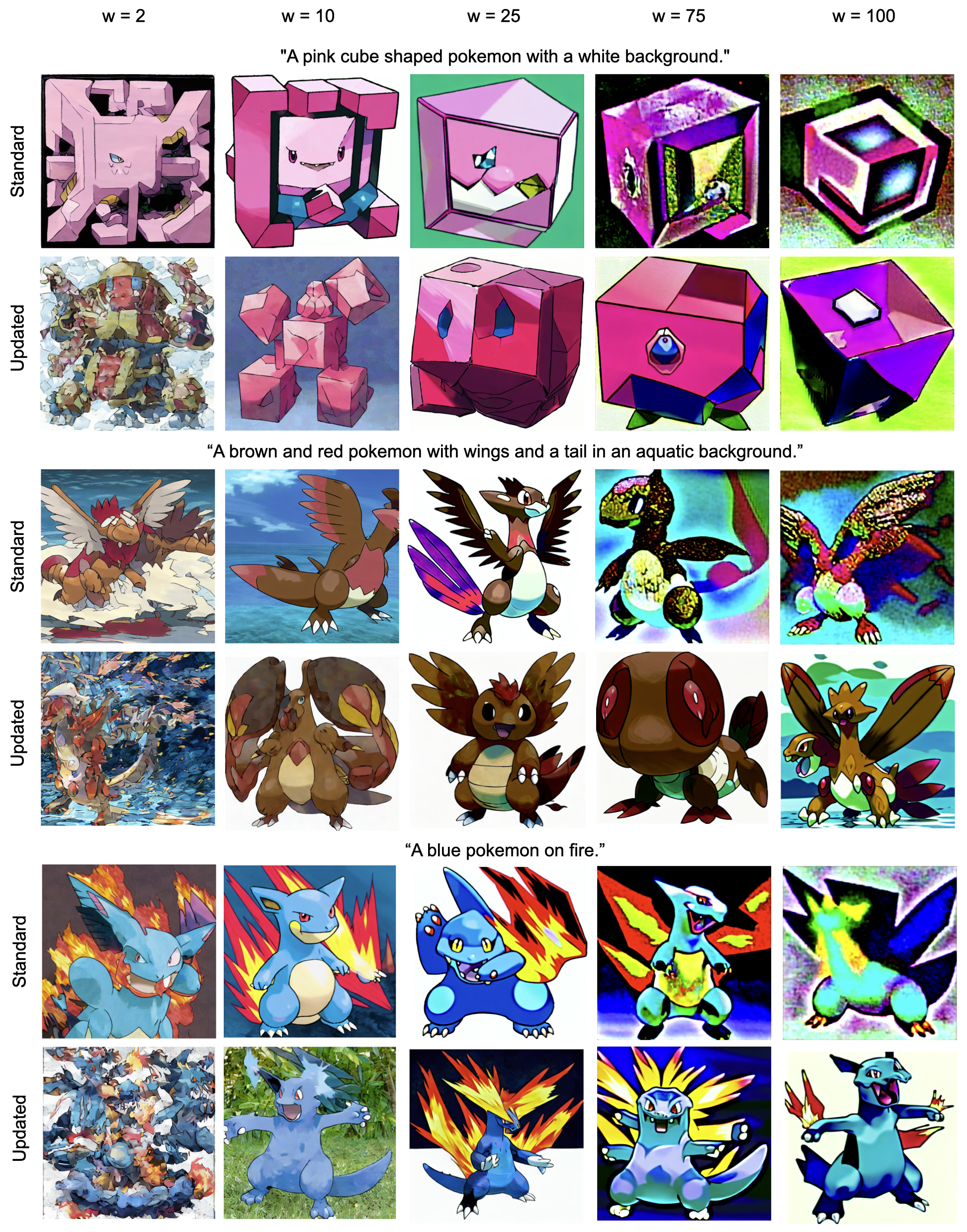}
  \caption{More samples comparing the standard c.f.g. guidance and the updated loss. These are all computed the same way as in ~\ref{fig: pokemon} (with 100 DDIM timesteps).}\label{fig: pokemon2}
\end{figure}

\clearpage
\subsection*{CIFAR-10 samples}

\begin{figure}[ht]
  \centering
  \includegraphics[width = 10.5 cm]{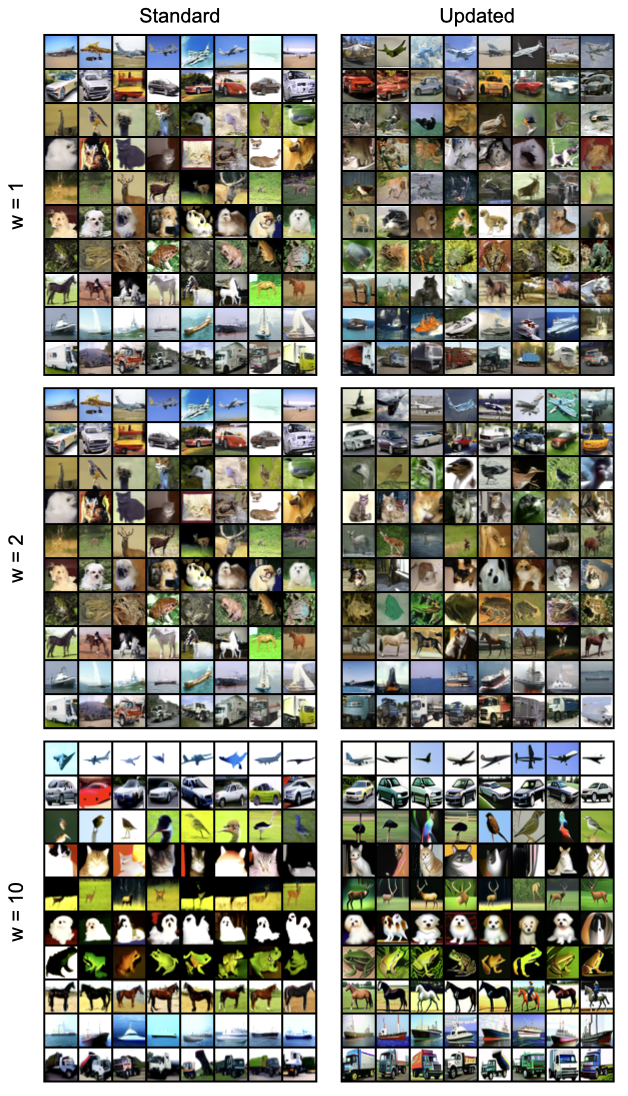}
  \caption{Example of samples generated with the respective models at \(w = 1.8\). Even though it is difficult to qualitatively assess these results, we can observe a more robust generation for the updated loss across all the values of $w$ used.}\label{fig: cifar2}
\end{figure}


\end{document}